\documentclass{egpubl}
\usepackage{eg2019}

\ConferenceSubmission   
 \electronicVersion 

\ifpdf \usepackage[pdftex]{graphicx} \pdfcompresslevel=9
\else \usepackage[dvips]{graphicx} \fi

\PrintedOrElectronic

\usepackage{t1enc,dfadobe}

\usepackage{egweblnk}
\usepackage{cite}

\usepackage{pifont}
\newcommand{\cmark}{\ding{51}}%

\usepackage{algpseudocode,algorithm, algorithmicx}

\newcommand*\Let[2]{\State #1 $\gets$ #2}
\algrenewcommand\algorithmicrequire{\textbf{Input:}}
\algrenewcommand\algorithmicensure{\textbf{Output:}}
\algnewcommand\algorithmicforeach{\textbf{for each}}
\algdef{S}[FOR]{ForEach}[1]{\algorithmicforeach\ #1\ \algorithmicdo}

\usepackage{booktabs} 
\usepackage{bm}
\usepackage{graphicx}
\usepackage{mathtools}
\usepackage{amsmath}
\usepackage{amssymb}
\usepackage{color}
\usepackage{mathtools}
\usepackage{multirow}
\usepackage{subcaption}
\usepackage{tikz}
\usetikzlibrary{bayesnet}
\usepackage{libertine}

\newcommand{\sect}[1]{Sec.~\ref{sec:#1}}

\newcommand{\fig}[1]{Fig.~\ref{fig:#1}}
\newcommand{\tab}[1]{Table~\ref{tab:#1}}

\newcommand{\given}{\,|\,}

\newcommand{\hdg}[1]{\vspace{0em}\noindent\textbf{#1}:}

\newcommand{\rv} [1]{\ensuremath{#1}}
\newcommand{\rt} {\rv{r}}

\newcommand{\nmd} {\ensuremath{\theta_{n.m.}}}
\newcommand{\cod} {\ensuremath{\theta_{c.o.}}}

\title[Automatic Generation of Constrained Furniture Layouts]%
      {Automatic Generation of Constrained Furniture Layouts}

\author[Paul Henderson, Kartic Subr \& Vittorio Ferrari]{
{\large Paul Henderson~\textsuperscript{1}, Kartic Subr~\textsuperscript{1}, and Vittorio Ferrari~\textsuperscript{1 2}}
\\[3pt]
{\small paul@pmh47.net ~~ ksubr@staffmail.ed.ac.uk ~~ vittoferrari@google.com} \\[4pt]
{\normalsize \textsuperscript{1} School of Informatics, University of Edinburgh ~~ \textsuperscript{2} Google Research, Z\"{u}rich}
}

\begin{document}

\teaser{
  \includegraphics[width=\linewidth]{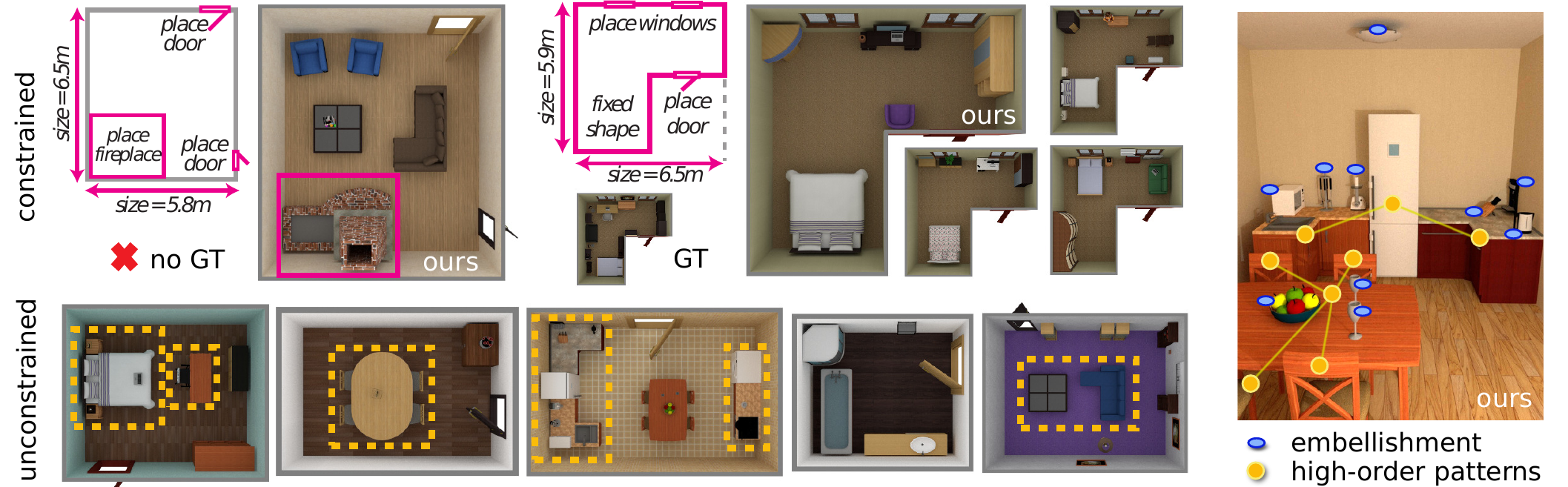}
  \centering
  \caption{
     We present a generative model for realistic furniture layouts which accommodates a variety of constraints such as room shape, size and object placement.
     Our model discovers high-order relations among objects (yellow annotations) and embellishes layouts (blue annotations) with items such as knives, laptops, fruit-baskets, etc.
     Our model is able to generate a valid layout in cases (top left) where there is no layout in the training set which satisfies the given size and placement constraints.
     We visualize four samples generated by our model for the L-shaped bedroom, along with the only human-designed room (GT) in the training set that satisfies the constraints.
     Our model supports efficient unconstrained sampling (0.04s per room). \emph{Figure best viewed on screen.}}
 \label{fig:teaser}
}

\maketitle
\begin{abstract}
Efficient authoring of vast virtual environments hinges on algorithms that are able to automatically generate content while also being controllable. 
We propose a method to automatically generate furniture layouts for indoor environments. 
Our method is simple, efficient, human-interpretable and amenable to a wide variety of constraints. 
We model the composition of rooms into classes of objects and learn joint (co-occurrence) statistics from a database of training layouts.
We generate new layouts by performing a sequence of conditional sampling steps, exploiting the statistics learned from the database. 
The generated layouts are specified as 3D object models, along with their positions and orientations.
We show they are of equivalent perceived quality to the training layouts, and compare favorably to a state-of-the-art method.
We incorporate constraints using a general mechanism -- rejection sampling -- which provides great flexibility at the cost of extra computation.
We demonstrate the versatility of our method by applying a wide variety of constraints relevant to real-world applications.


\end{abstract}  

\section{Introduction}

Large scale virtual environments are an important feature across a plethora of applications such as massively-multiplayer-online computer games, movies, and even simulators used to train self-driving cars.
Manual authoring of environments at these scales is impractical and there is a need for algorithms that can automatically generate realistic virtual environments. 
To be considered useful, it is important that automatic algorithms are able to accommodate constraints stipulated by domain experts.
In this paper, we address the problem of automatically generating \textit{furniture layouts} for indoor scenes. 
We propose a simple algorithm to automatically populate empty rooms with realistic arrangements of 3D models of furniture and other objects, using occurrence and placement statistics learned from a training database of layouts of CAD models.
Our method supports a wide variety of \textit{a priori} constraints.
It can be used in conjunction with methods such as \cite{merrell10tog,Ma:2014:GLL} that place rooms within buildings, to automatically generate complete indoor virtual environments.

The computer graphics literature is rich with methods that enable 3D content creation, from landscapes~\cite{Cordonnier:2017} and urban sprawls~\cite{Parish:2001} to individual objects such as furniture~\cite{li_sig17}, buildings~\cite{Muller:2006,Nishida:2016}, etc. 
These methods involve varying degrees of user-interaction to achieve realism and/or aesthetic appeal.
Procedural approaches rely on parametric controllability while methods that are posed as optimization rely on constraint-driven controllability. 
A third class of methods adopts a \textit{data-driven} approach to generate or edit models based on features learned from training examples~\cite{funkhouser2004modeling,Emilien:2015,li_sig17}. 

We propose a new data-driven, probabilistic, generative model for 3D room layouts (\sect{algorithm}).
Our model learns statistics from a database~\cite{song17cvpr} containing over $250\,000$ rooms designed by humans (\sect{training}).
We categorize over $2500$ 3D models from the database and learn conditional statistics across these categories.
The output space of this learned model is a high-dimensional combination of discrete and continuous variables.
Our model is based on directed, acyclic dependencies between categories, which allows easy and very efficient ancestral sampling.
Each sample results in a furniture layout.
We demonstrate the flexibility of this approach by incorporating user-specified constraints such as accessibility, sizes and shapes of rooms, locations of doors and windows, and constraints on the location of furniture items such as sofas and television screens (\sect{constraints}).
Finally, we present a user study showing that layouts generated by our model are realistic (equal quality to those in the training set), and compare favorably to the state-of-the-art method \cite{wang18tog} (\sect{results}).
%
%


\subsection{Related work}

\begin{table}
  \begin{center}
    \scalebox{.85}
    {

    \begin{tabular}{r|cccccccc}
      &
      [1] & [2] & [3]& [4] & [5] &[6] &[7] & ours
      \\
      \hline
      room type
      & \cmark
      & \cmark
      & -
      & \cmark
      & \cmark
      & -
      & -
      & \cmark
      \\
      room size
      & \cmark
      & \cmark
      & \cmark
      & \cmark
      & \cmark
      & \cmark
      & \cmark
      & \cmark
      \\
      room shape
      & \cmark
      & -
      & -
      & -
      & -
      & -
      & -
      & \cmark
      \\
      traversability
      & -
      & -
      & -
      & -
      & -
      & \cmark
      & \cmark
      & \cmark
      \\
      object existence
      & -
      & -
      & \cmark
      & -
      & -
      & \cmark
      & \cmark
      & \cmark
      \\
      object placement
      & -
      & -
      & -
      & -
      & -
      & -
      & \cmark
      & \cmark
      \\ \hline
      fully automatic
      & \cmark
      & \cmark
      & -
      & \cmark
      & \cmark
      & -
      & -
      & \cmark
    \end{tabular}
    }
    \caption{\label{tab:relworktable} Comparison of constraints supported by prior methods and ours. Legend: 1 =\cite{wang18tog}; 2 =\cite{qi18cvpr}; 3 =\cite{Fu:2017}; 4 =\cite{Liang:17};  5 =\cite{sadeghipourkermani:2016:learnsynth}; 6 =\cite{yu11siggraph}; 7 =\cite{merrell11siggraph}. `Fully automatic' methods support generation without user specification of any furniture classes.}
  \end{center}
\end{table}

\hdg{Priors for understanding structure in indoor scenes}
%
%
Choi et al~\shortcite{choi13cvpr} performed 3D object detection and layout estimation by inferring spatial relations between objects using a discriminative, energy-based formulation.
They do not present a generative model over layouts.
%
Zhao and Zhu~\shortcite{zhao11nips,zhao13cvpr} built a probabilistic grammar model, using specifically engineered production rules, over cube-based 3D structures constituting parts of rooms.
This grammar generates arrangements of coarse blocks and does not result in layouts of entire rooms.
Similarly, treating objects as cuboids, Del Pero et al~\shortcite{delpero12cvpr} proposed a generative model over room size and layout using learned distributions for the dimensions of the cuboids.
The model does not learn inter-object relationships such as co-occurrence or relative locations. Although the importance of such relationships was discussed in follow-up work~\cite{delpero13cvpr}, inter-object relationships were not incorporated in the generative model.

\begin{figure}
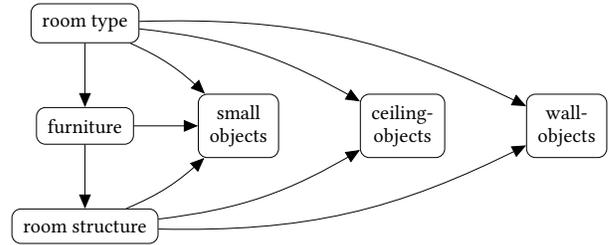

  \begin{center}
\scalebox{.85}
{
  \tikz[box/.style={rectangle, rounded corners, draw=black, font=\fontsize{9}{10}\selectfont, align=center, inner sep=5pt}]{
  \node[box ] (room-type) {room type} ;
  \node[box, below=of room-type] (furniture) {furniture};
  \node[box, below=of furniture] (layout) {room structure};
  \node[box, right=of furniture] (small-objects) {small\\objects} ;
  \node[box, right=36pt of small-objects] (ceiling-objects) {ceiling-\\objects} ;
  \node[box, right=36pt of ceiling-objects] (wall-objects) {wall-\\objects} ;
  \edge {room-type} {furniture} ;
  \edge[bend left=10] {room-type} {small-objects} ;
  \edge[bend left=12] {room-type} {ceiling-objects} ;
  \edge[bend left=14] {room-type} {wall-objects} ;
  \edge {furniture} {layout,  small-objects} ;
  \edge[bend right=10] {layout} {small-objects} ;
  \edge[bend right=12] {layout} {ceiling-objects} ;
  \edge[bend right=14] {layout} {wall-objects} ;
  }
}
  \caption{\label{fig:model-structure} Elements of our generative model with arrows representing conditional dependencies.
  The directional dependencies enable \textit{ancestral sampling}.
  Given a sample $\rv{Q_o}\sim p_o(x)$, generated at node \rv{o}, we can draw a sample from its child node, \rv{u}, according to $Q_u \sim p_u(x \given \rv Q_o)$.
  Different categories of object (middle row) are sampled conditional on
  the room type; the room structure (i.e. sizes of the cells) is defined by the furniture it
  contains.
  }
  \end{center}
\end{figure}

\hdg{Generating realistic furniture layouts}
A common approach is to adopt an energy-based formulation~\cite{merrell11siggraph,yu11siggraph,sadeghipourkermani:2016:learnsynth,handa16icra,qi18cvpr} with potentials between objects to impose constraints and preferences.
The method of Handa et al~\shortcite{handa16icra} generates room layouts by optimizing a pairwise energy term using simulated annealing, with random initialization.
Sadeghipour Kermani et al~\shortcite{sadeghipourkermani:2016:learnsynth} propose a method for generating bedrooms, that separates the sampling of classes and counts from the spatial arrangement.
Liang et al~\shortcite{Liang:17} also propose a two-step method, demonstrated on five room types.
Objects are first selected based on statistics learned from a database, without considering inter-object relationships.
Their locations are then chosen by MCMC sampling from a pairwise energy model.
None of these methods reason about smaller objects positioned on the furniture, nor objects mounted on the walls or ceiling.
Moreover, both simulated annealing and MCMC are slow, and not guaranteed to converge to a valid layout.

Very recently, Qi et al~\shortcite{qi18cvpr} represent indoor scenes using a probabilistic grammar, using a conditional Gibbs distribution as prior on its parse graphs. The conditioning parameter is learned from a large database.
Their approach requires considerable manual modeling including specification of potential functions and grouping relations between objects  such as chairs and tables.
Novel layouts are generated using MCMC sampling, along with simulated annealing; this takes around 2400s per layout.
Wang et al~\shortcite{wang18tog} exploit the power of convolutional neural networks (CNNs). Their method generates layouts by using three different CNNs to decide whether to add furniture, what furniture to add, and where to place it.
This approach avoids a costly MCMC optimization process, but still takes several minutes on a GPU to sample a single room, and several days to train the models.
Both these approaches are trained with the same dataset that we use in this paper.
While they are fully automatic, neither supports inputing user-specified constraints.

Other methods suggest new layouts based on partial input, such as the object classes and an initial arrangement~\cite{merrell11siggraph,yu11siggraph,Fu:2017}.
Fu et al~\shortcite{Fu:2017} exploit knowledge of how human agents interact with indoor environments to synthesize entire room layouts from a room shape and a few classes of objects that are to be present.
This method is the fastest among prior works, taking approximately 2s to generate a layout.

Finally, some works model small areas of rooms centered around one item of furniture, instead of complete layouts.
Fisher et al~\shortcite{fisher12siggraph} propose a method that learns from a dataset of 130 layouts, to embellish parts of rooms by adding relevant objects around (e.g. chairs around a table) as well as on (e.g. books on a table) furniture. It does not model entire layouts of rooms containing complex arrangements of furniture.
Ma et al~\shortcite{ma16tog} again generate small areas of rooms by embellishing a focal object, but they decide the classes and locations of objects to add by reasoning over human actions that could take place in the scene.




\begin{figure}
  \begin{center}
    \includegraphics[width=0.95\linewidth]{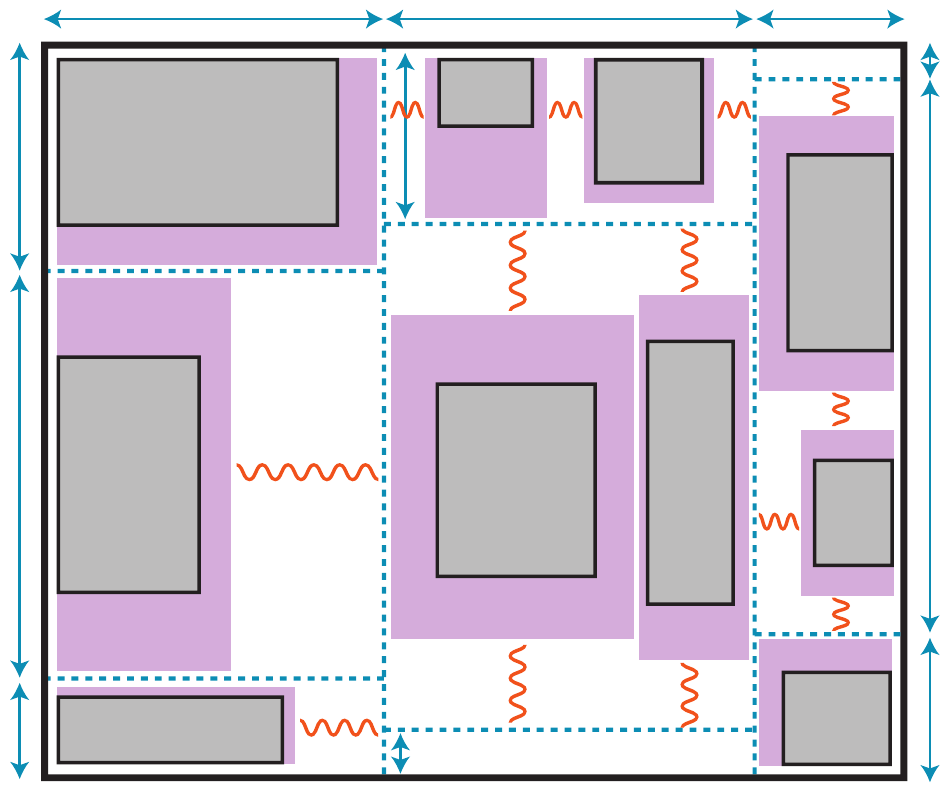}
    \caption{ \label{fig:cellstructure} Cell structure for furniture layout. Thick black lines represent the walls enclosing the room, gray boxes represent furniture objects, dashed blue lines delimit cells and blue arrows indicate dimensions that expand/contract to fit their contents. There is one cell along each edge of the room, one at each corner, and for larger rooms, one in the interior. Other cell structures can capture prior knowledge about the functional layout of specific room types, e.g.~a conversation area in a living room. Objects are padded with free space on each side (purple). Objects in cells that are larger than the sum of their contents, due to constraints due to neighboring cells, are distributed evenly (orange springs), with the exception that objects in corner/edge cells always remain flush with walls.
    }
  \end{center}
\end{figure}

\hdg{Summary and motivation}
Although many algorithms have been proposed to generate furniture layouts, none of them is fully automatic, considers inter-object relationships, and is amenable to a diverse range of user-specified constraints (\tab{relworktable}).
Most automatic methods are limited to simple constraints such as room type, shape, size and the presence of a particular object.
While other constraints could be applied by rejection sampling, this is prohibitively expensive, as these models are at best $50\times$ slower than ours to sample from, and even slower for methods requiring no user interaction.
Adding constraints to energy-based models is possible by introducing new potentials, but this requires careful design and engineering.
Moreover, it would make sampling even more costly.
Thus, it is not practical to incorporate complex constraints in any previous approaches to layout generation.

\subsection{Contributions} \label{sec:contrib}
We make the following contributions:
\begin{enumerate}
\itemsep=0em
\listparindent=0mm
\itemindent=0mm
\leftmargin=0em
\labelsep=.5em
 \item We propose a generative model for room layouts, based on a directed graphical model that captures high-order relationships between objects;
 \item Our generative model enables efficient, ancestral sampling of objects and their attributes; and
 \item Our method is fully automatic and allows specification of general \textit{a priori} constraints using rejection sampling.
\end{enumerate}

\section{Automatic generation of room layouts} \label{sec:algorithm}


%
%
We begin by sampling a \textit{room type} (e.g.~kitchen, living room), then sequentially sample furniture instances, conditioned on the room type and instances already sampled.
We partition rooms into \textit{cells} and sample objects and their positions within these cells (Fig.~\ref{fig:cellstructure}) such that geometric intersections will not occur.
We begin with furniture objects (Sec.~\ref{sec:corealgo}), which define the structure of the room.
Then, in an \textit{embellishment} step (Sec.~\ref{sec:embellish}), we sample instances of other categories, given the furniture items and their locations.
The overall sampling process is illustrated in \fig{model-structure}.

The parameters used for sampling instances of objects, for cell assignments and positioning within a cell are learned from the dataset (Sec.~\ref{sec:training}).
These are simple parametric models, which are human-interpretable and modifiable.

\begin{figure}
    \begin{center}
        \includegraphics[width=0.90\linewidth]{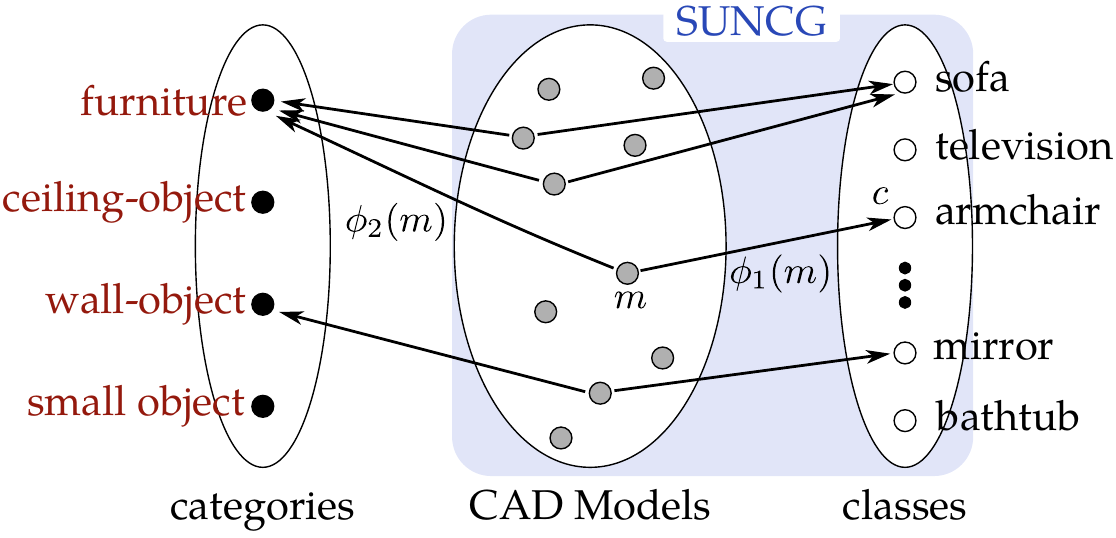}
        \caption{\label{fig:categories} We introduce a set of \textit{categories} (left) along with a manually-specified mapping $\phi_2$ from models to categories. The SUNCG database already specifies a mapping $\phi_1$ from models to object classes. We use both mappings.}
    \end{center}
\end{figure}

\subsection{Classes and categorization} \label{sec:strcateg}
The dataset contains CAD models, where each model $m$ is assigned an \textit{object class} $\phi_1(m)$ such as ``television'', ``bathtub'', ``armchair'', etc.
There is a total of 2500 CAD models and 170 object classes in the collection.
We introduce an additional set of labels called \textit{categories}: furniture, small objects such as books and laptops, wall objects such as picture frames, and ceiling objects such as lamps. We manually specified a second mapping $\phi_2$ from models to categories (Fig.~\ref{fig:categories}).
Our sampling strategy for models differs between the different categories, as discussed in the next sections.

\subsection{Sampling furniture} \label{sec:corealgo}
We place furniture instances by sampling counts of \textit{singletons} (i.e.~individual objects), and of \textit{motifs} and \textit{abutments}, which are spatially-related groups learned from the training data.
For each instance, we sample a cell it is to be placed in (e.g.~`against the east wall'), and its orientation and padding.
After all furniture counts and instance parameters have been sampled, we position the resulting objects deterministically.
The \textit{room structure} is finalized when cells (and hence the room) are sized to accommodate the objects and their paddings.

\hdg{Singletons}
We sample singletons using algorithm~\ref{alg:sampfurniture},
where functions SampleNumInstances, SampleCell, SampleOrientation and SamplePadding sample from distributions whose parameters are learned from training data. 
The arguments to these functions signify what the underlying distributions are conditional on.
At each iteration (line 6 of algorithm~\ref{alg:sampfurniture}), cells expand to fit the sampled objects, ensuring no intersections between objects.

\begin{algorithm}[t]
  \caption{Sampling singleton furniture instances}
    \label{alg:sampfurniture}
  \begin{algorithmic}[1]
    \Require{\rt\ is the room type
    }
    \Statex
    \Function{SampleFurniture}{\rt}
      \ForEach{object class $c$} 
	\Let{$n_c$}{$0$}
	\ForEach{model $m$ \textbf{with} $(\phi_1(m) = c) \wedge (\phi_2(m) = \mathit{furniture})$}
          \Let{$n_m$}{SampleNumInstances($m$, $n_c$, $r$)}
	  \For{$j \gets 1 \textrm{ to } n_m$} \Comment{$n_m$ instances of $m$}
	    \Let{$k_j$}{SampleCell($m$)}
	    \Let{$\theta_j$}{SampleOrientation($m$, $k$)}
	    \Let{$p_j$}{SamplePadding($m$)}	    
          \EndFor
	  \Let{$n_c$}{$n_c+n_m$} \Comment{accumulate count}
	\EndFor
      \EndFor
    \EndFunction
    \Statex
      \Ensure{for each CAD model $m$:}
      \Statex{\hspace{0.28cm} (i) number of instances $n_m$ that we place;}
	  \Statex{\hspace{0.20cm} (ii) parameters $\left\{k_{j},\, \theta_{j},\, p_{j}\right\}_{j=1}^{n_m}$ of each instance of $m$. }
  \end{algorithmic}
\end{algorithm}


\hdg{Motifs} 
Motifs are groups of items that are present together in many examples of the dataset, such as a table with chairs around it.
We sample counts and instance parameters for each motif following lines 5--10 of algorithm~\ref{alg:sampfurniture}, but in this case $m$ represents a motif rather than a singleton CAD model.
Then, we set the relative offsets and orientations between items within a motif as observed in a randomly selected instance of the motif in the training database.
This non-parametric strategy for determining relative placement yields coherent, visually-pleasing groups of furniture, and eliminates the need for expensive collision/intersection tests between meshes.

\hdg{Abutments}
Abutments are groups of items that appear in rows, abutting one another, with variations in their sequence,~e.g.~a row of cabinets in a kitchen along with a dishwasher, refrigerator and/or washing machine.
Again, we sample counts and instance parameters for each abutment using lines 5--10 of algorithm~\ref{alg:sampfurniture}, where $m$ now represents a class of abutment.
The furniture items within an abutment are modeled as a Markov chain with a terminal state; for each instance of the abutment, we sample from this Markov chain to obtain a specific sequence and number of CAD models. 
The transition probabilities of the Markov chains are learned during training.

\subsection{Embellishment} \label{sec:embellish}

\hdg{Ceiling objects}
Given a room type \rt, we draw a single CAD model $m$ randomly according to a discrete probability mass function (pmf) \cod\ over models in this category.
The number of instances of $m$ is determined using $\mathrm{SampleNumInstances}(m,0)$.
This number is rounded up so that it can be factorized into a product of integers and the objects are positioned on a grid.

\hdg{Wall objects}
For each CAD model $m$ that is a wall object,
we determine the number of instances in similar fashion to furniture in lines 2--5 of algorithm~\ref{alg:sampfurniture}.
Each instance is then assigned to a wall uniformly randomly and its position on the wall is a combination of a Normal distribution along the $Y$ axis and Uniform distributions in $X$ and $Z$.
The parameters of the Normal distribution are learned (conditioned on $m$). 
If this results in a geometric conflict (intersection with other wall objects, doors, etc.) we reject the sampled location and repeat the process until there are no conflicts.

\hdg{Small objects}
For each furniture instance with CAD model $m$, we sample small objects non-parametrically conditioned on $m$ and \rt. We choose a random instance of $m$ in a room of type \rt\ from the database, and replicate the configuration of small objects associated with that instance.

\subsection{Algorithm summary}
To summarize, we first randomly sample the type of room \rt\ from a discrete distribution over 9 room types found in the database. 
The distribution (pmf) of \rt\ is learned during training. Then, we sample furniture items conditioned on \rt: first singletons, then motifs and finally abutments.
The numbers and instances of each item are determined by parameters learned during training.
Once all furniture items are sampled, and assigned to cells, we use a deterministic placement algorithm that calculates their final positions in the room. 
Then, we sample ceiling objects and wall objects conditioned on \rt\ and the furniture placed. Finally, we sample small objects conditional on the furniture in the room and \rt{}.

\section{Training} \label{sec:training}

\begin{figure}
    \centering
    \includegraphics[width=0.48\linewidth]{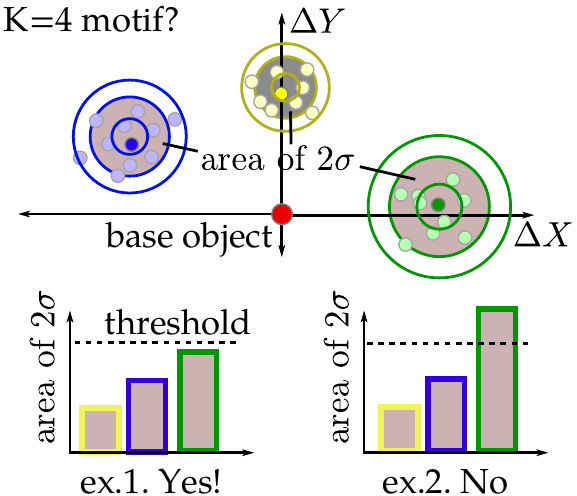}
    \includegraphics[width=0.48\linewidth]{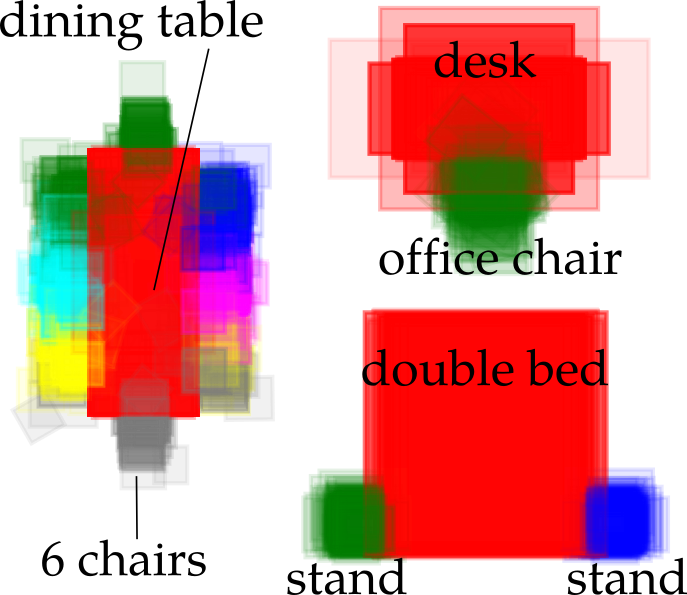}
    \caption{\label{fig:example-motifs} Motif discovery. \emph{Left:} We use DPMM clustering for K tuples of object occurrences and identify motifs as those tuples for which displacements from a base object in the tuple is within some threshold. \emph{Right:} Examples of motifs that we automatically discover in SUNCG.
    Each colour corresponds to a different object in the pattern; we overlay 200 occurrences of each pattern to illustrate its variability.
    The red objects are the base objects of the patterns.}
\end{figure}

\hdg{Dataset}
We use a large dataset of ground-truth room layouts to learn parameters that are then used for automatically generating layouts.
SUNCG~\cite{song17cvpr} is a new dataset of 45000 apartment layouts, created by humans, and separately verified as reasonable by humans.
Each apartment has an average of 8.1 rooms; the majority are annotated with the room type.
The apartments are designed with an online tool, and contain objects of 170 classes, represented by around 2500 CAD models.
There are 4.5M object instances; each consists of a reference to a CAD model, and its location and orientation in the room.

\hdg{Number of instances} 
We model the number of instances $n_m$ of each CAD model as being conditional on the model $m$, the room type $r$, and on the number $n_c$ of furniture instances already sampled for the class $\phi_1(m)$.
The distribution (pmf) \nmd\ over count bins $\{0,1,2,3,4, >4\}$ is calculated as a histogram (normalized) over all scenes in the database of type $r$. 
Further, a Poisson distribution is fitted to the observed $n_m$ in all scenes with $n_m > 4$.
SampleNumInstances (line 5 of algorithm~\ref{alg:sampfurniture}) is implemented in two steps.
First, we draw an indicator variable according to \nmd. If this variable is less than or equal to 4, then we return it as the number of instances. 
Otherwise, we return a sample from the Poisson distribution.

\hdg{Instance attributes}
For each model $m$, during training, we calculate a pmf over 9 cells (4 corners, 4 edges and internal) by normalizing the histogram of occurrences.
We implement SampleCell (line 7 of algorithm~\ref{alg:sampfurniture}) by returning a cell according to this pmf.
For models in internal cells, we count the number occurrences where they are aligned (positively or negatively) with respect to any axis and the number of ``non-aligned'' instances, and use this to learn a pmf.
We implement SampleOrientation (line 8 of algorithm~\ref{alg:sampfurniture}) by sampling an indicator variable according to the pmf for orientations.
If this variable indicates non-alignment, we sample an orientation uniformly at random.
Finally, we model padding around CAD models as a 4D diagonal-covariance Normal distribution conditioned on the CAD model $m$.
The dimensions correspond to padding on each side of the object: in-front-of, behind, to-the-left-of and to-the-right-of.
SamplePadding (line 9 of algorithm~\ref{alg:sampfurniture}) returns a sample from this 4D Normal distribution.
The knowledge learnt during training is interpretable; \fig{interpretable}b,c show values from the pmfs, indicating typical placements of objects, while \fig{interpretable}a shows typical locations where we place various classes.
In both cases, these agree well with human intuition on interior design.


\begin{figure}
    \centering
    \includegraphics[width=\linewidth]{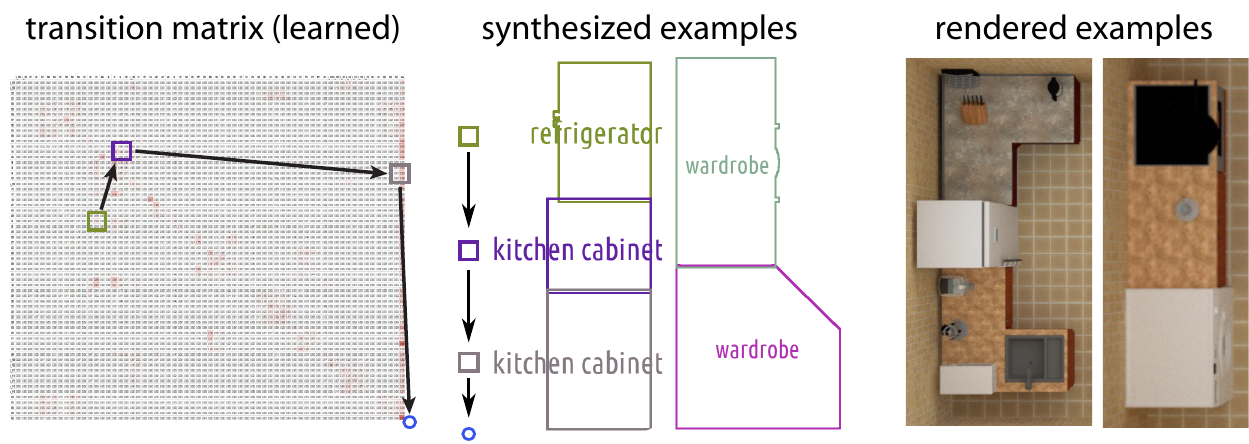}
    \caption{Statistics of abutments observed in the training database are recorded in a transition matrix, along with a terminal state (blue circle). We synthesize abutments by generating Markov chains using the learned transition probabilities.}
    \label{fig:example-abutments}
\end{figure}

\hdg{Motif discovery}
We search the training set for all joint occurrences of a given $K$-tuple of classes (e.g.~table, chair, chair) within a room.
For every occurrence of one of these classes -- designated as the \textit{base} object -- we calculate displacements of the centres of the other objects relative to the base object.
We model these displacements as points in a $2(K-1)$-dimensional space and cluster them with a Dirichlet process mixture model (DPMM)~\cite{rasmussen00nips}, fitted by variational inference. 
We use Gaussian clusters with diagonal covariance and fit a DPMM per $K$-tuple of classes. 
We calculate the area inside the $2\sigma$ contour for the location of each element in the motif; if all of these are less than a threshold, then the cluster is accepted as a motif.
We store the CAD models, relative locations, and orientations for every occurrence assigned to the cluster; one of these will be selected when instantiating the pattern.
Some examples of motifs we discover are given in Fig.~\ref{fig:example-motifs}.

\hdg{Abutment discovery}
We discover abutments in two stages.
First, we gather sets $\mathcal{S}_i$ of sequences of CAD models, where each set will ultimately become an abutment pattern.
Each sequence of CAD models represents an abutting series of instances in the training set.
Then, for each $\mathcal{S}_i$, we calculate the transition probabilities for the corresponding Markov chain, as maximum-likelihood estimates given the CAD model sequences $\mathbf{s} \in \mathcal{S}_i$.
More precisely, we collect such sets $\mathcal{S}_i$ in a collection $\mathcal{T}$, initialising $\mathcal{T}$ to be empty.
While doing so, we maintain the invariant that $\forall i \neq j$, $\mathcal{S}_i$ and $\mathcal{S}_j$ do not contain any sequences that share CAD models.
For each room in the training set, we find all pairs of objects that abut, based on their rotated bounding-boxes touching at an edge.
These pairs are combined transitively to form full sequences of objects $\mathbf{s}_j$, each being a row of abutting objects of some orientation.
For each object-sequence $\mathbf{s}_j$, ignoring those with just two objects, we check if any of its CAD models appears in any sequence in a set $\mathcal{S}_i \in \mathcal{T}$ already created.
If so, we add the object-sequence to $\mathcal{S}_i$; if not, we create a new one storing just $\mathbf{s}_j$, and add it to $\mathcal{T}$.
In the first case, we also check that adding the sequence to $\mathcal{S}_i$ has not broken the invariant that sets do not share CAD models; if it has, we merge sets until the invariant holds again.
At the end of the above process, each $\mathcal{S}_i \in \mathcal{T}$ contains many sequences of CAD models, each of which we will treat as a sample from the Markov chain $M_i$.
It is then straightforward to learn the transition probabilities for $M_i$ by maximising the likelihood of all the sequences $\mathbf{s}_j \in \mathcal{S}_i$.
Some examples of abutments we discover are given in Fig.~\ref{fig:example-abutments}.

\section{Constraints} \label{sec:constraints}

Our generative model accommodates diverse constraints using rejection sampling as a general mechanism,~i.e.~we sample layouts until we obtain one that satisfies all constraints.
We demonstrate the versatility of our generative model using some example constraints.
Incorporating other constraints can be achieved similarly as long as a given layout can be verified to satisfy them.
Since our sampling process is very fast (tens of milliseconds per room), any inefficiency due to rejection sampling is outweighed by its ability to serve as a common mechanism to impose a wide range of constraints (Tab.~\ref{tab:rejection-rates}).
In some special cases, we can avoid rejection sampling by allowing users to explicitly manipulate parameters of distributions learned (Fig.~\ref{fig:custom-padding-samples}).

%

\hdg{Room type and size}
As the room type \rt\ has no ancestors in our model, it can directly be assigned a constrained value, avoiding rejection sampling entirely.
Since room size is a continuous value, and the probability of any sample satisfying this is zero, we allow a small tolerance on each dimension (2\% in all our examples).

\hdg{Traversability}
A layout is traversable if there exists a path between all points in free space (regions with no furniture) and from all points in free space to all doors in the room. 
To verify this, we first rasterise an orthographic projection of the furniture onto the floor plane at a fixed resolution and identify free space as the complement of this footprint.
We calculate $P$ (areas where people can stand or pass) via morphological erosion of the free space using a circular kernel of radius 0.25m and also add regions on doors to $P$.
Similarly we calculate regions $A$ that require access using morphological erosion, but with a larger kernel.
Then we verify traversability by checking whether $x$ is reachable from $y, \; \forall x,y\in A$ via some $\{z\} \subseteq P$.
%

\hdg{Gap placement and clearance}
Ensuring there is a gap at a particular location allows users to augment layouts with their own 3D models, that are not part of our system.
In order to make rejection sampling efficient, rather than just discarding layouts until one that satisfies the constraint is found, we directly place a `gap instance' in a suitable cell, ensuring that no object will occupy the relevant space.
Note that some rejections will still occur, as cell locations are not known precisely until all furniture items are placed.
We also allow users to specify the clearance around particular objects, regardless of where in the room they are placed, e.g.~to allow additional space next to a bed that must be used by a mobility-impaired individual.
This is implemented efficiently by directly adjusting the parameters of the relevant padding distribution.

\hdg{Object placement}
We allow users to place instances of CAD models known to our system, at specific locations -- e.g.~a bed against a particular wall.
Similarly to placing gaps, we ensure that a suitable instance is placed in the relevant cell, thereby greatly reducing the chances of rejection.

\hdg{Doors, windows and refurnishing}
We model door and window specification using a combination of gap-placement, at edges of rooms, and traversability (the relevant area is included in $P$).
This allows us to support \textit{refurnishing} existing rooms from SUNCG -- that is, generating new furniture, while retaining the existing size/shape, doors, and windows.
This is valuable for generating complete, realistic rooms without any user input. 

\section{Results} \label{sec:results}

In this section, we present qualitative results to highlight the samples (with and without constraints) generated using our model, and quantitative results measuring performance.
We also assess the quality of our generated layouts via a simple user study. 
All rendered images were obtained using path-tracing~\cite{Mitsuba}.
The execution times reported in this paper were obtained using our unoptimized and sequential Python implementation, on an Intel Xeon E5-2620v3 processor, using less than 1GB of RAM.

\begin{figure}
    \centering
    \includegraphics[height=0.92\linewidth, angle=90]{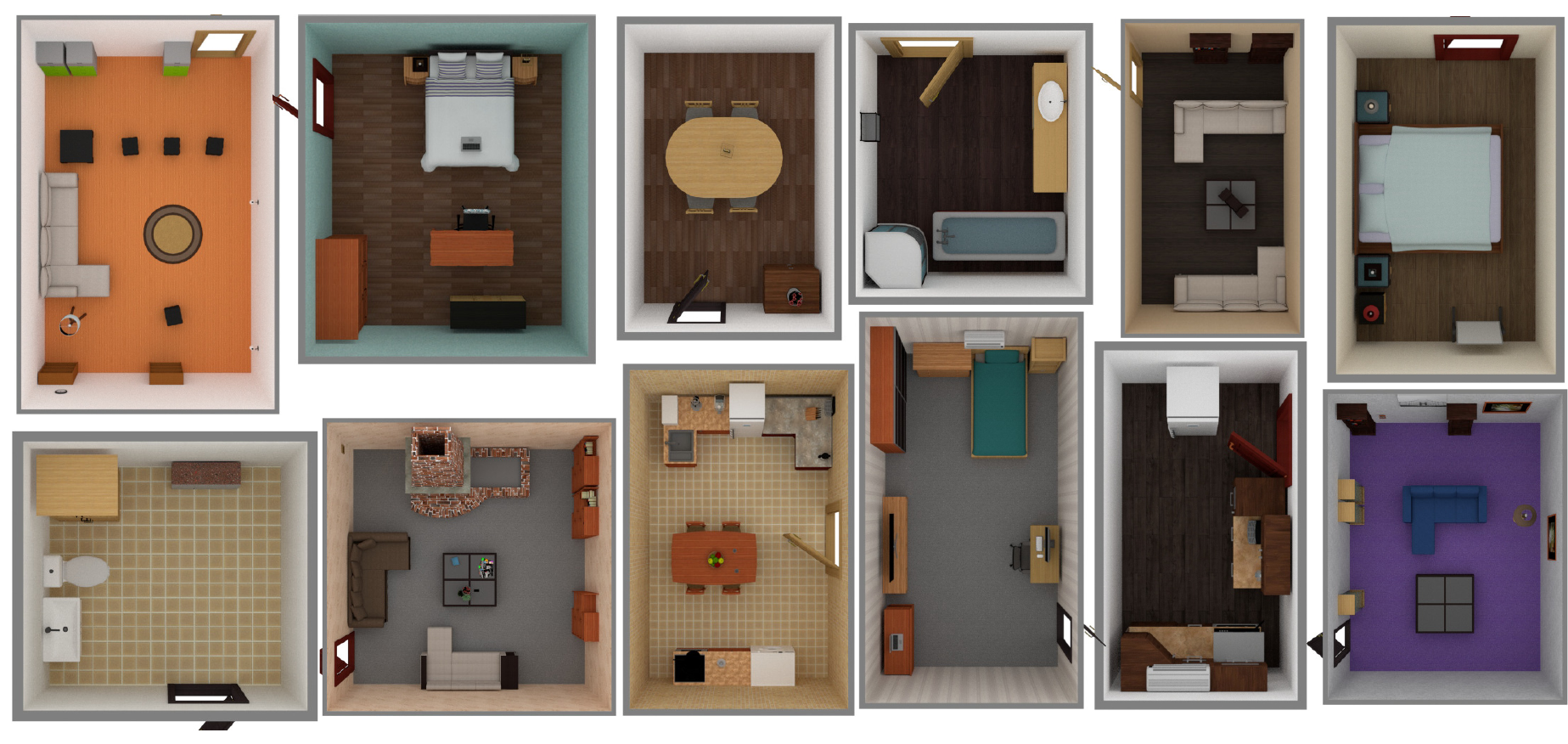}
    \vspace{-10pt}
    \caption{Samples from our model, without constraints applied}
    \label{fig:unconstr-samples}
\end{figure}

\subsection{Generating layouts}

\hdg{Unconstrained output}
In Figure~\ref{fig:unconstr-samples}, we show some output examples from our generative model, without any constraints specified. Our model produce results without objects intersecting and mimics the diversity found in the training dataset -- both in terms of the types of rooms as well as the objects in them.
The co-occurrence and relative placements of objects are also realistic and natural. Unconstrained layout samples are generated in 0.04s on average.
%

\begin{figure}
    \centering
      \includegraphics[height=\linewidth, angle=90]{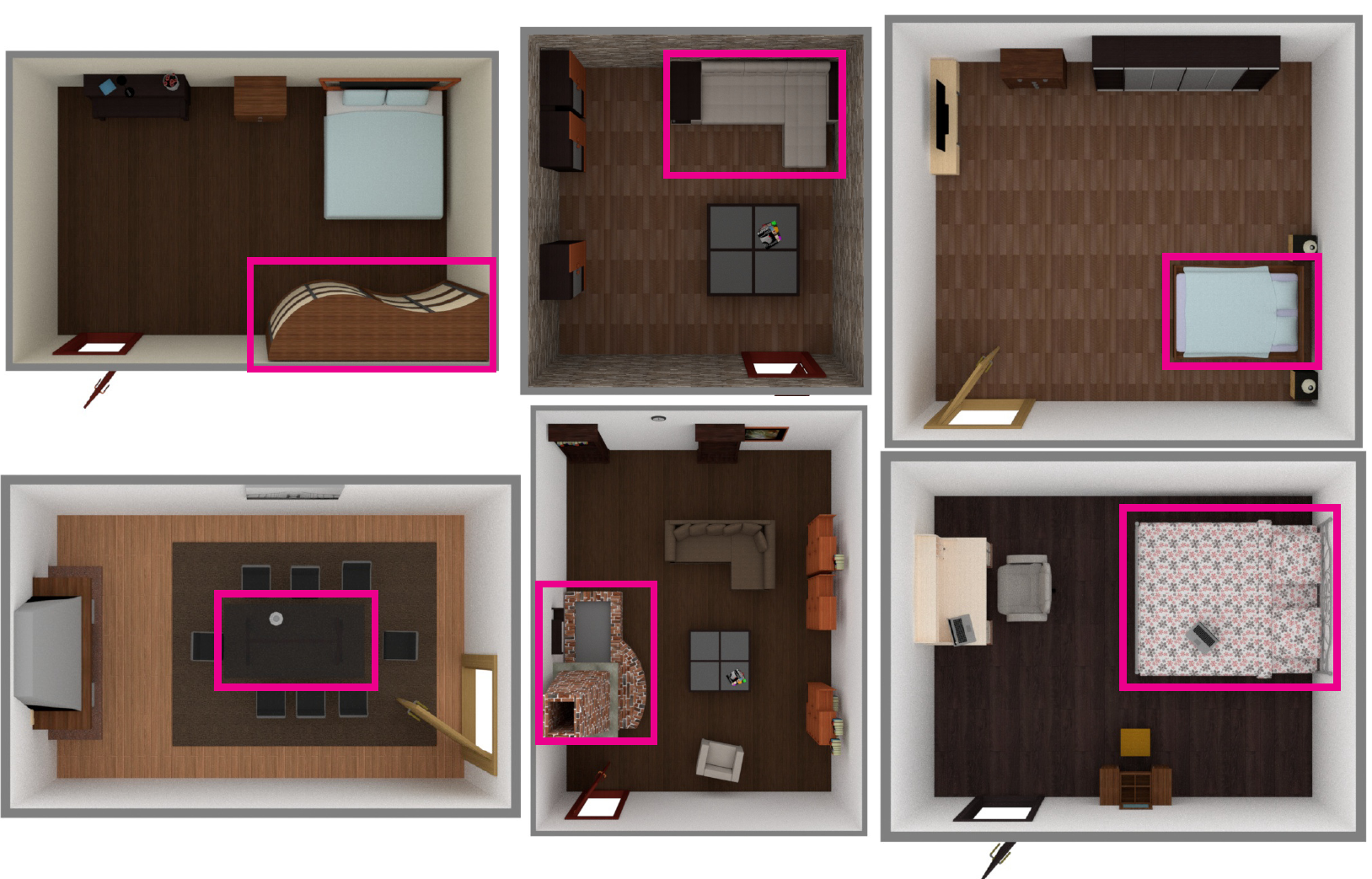}
    \caption{\label{fig:constr-samples-object} Samples from our model, with constraints. The sizes of the rooms and the locations/classes of objects shown in pink boxes are constrained}    
\end{figure}

\hdg{Examples with constraints}
Figure~\ref{fig:constr-samples-object} shows examples of layouts where the room size and placement of one object was specified by the user. Note that the other sampled objects in the room are automatically chosen, and placed harmoniously. For example, in the first image, the constraint was ``place a bed near the top right corner''. Our method automatically places nightstands on either side of the bed. 

\begin{figure}
    \centering
      \includegraphics[height=0.95\linewidth, angle=90]{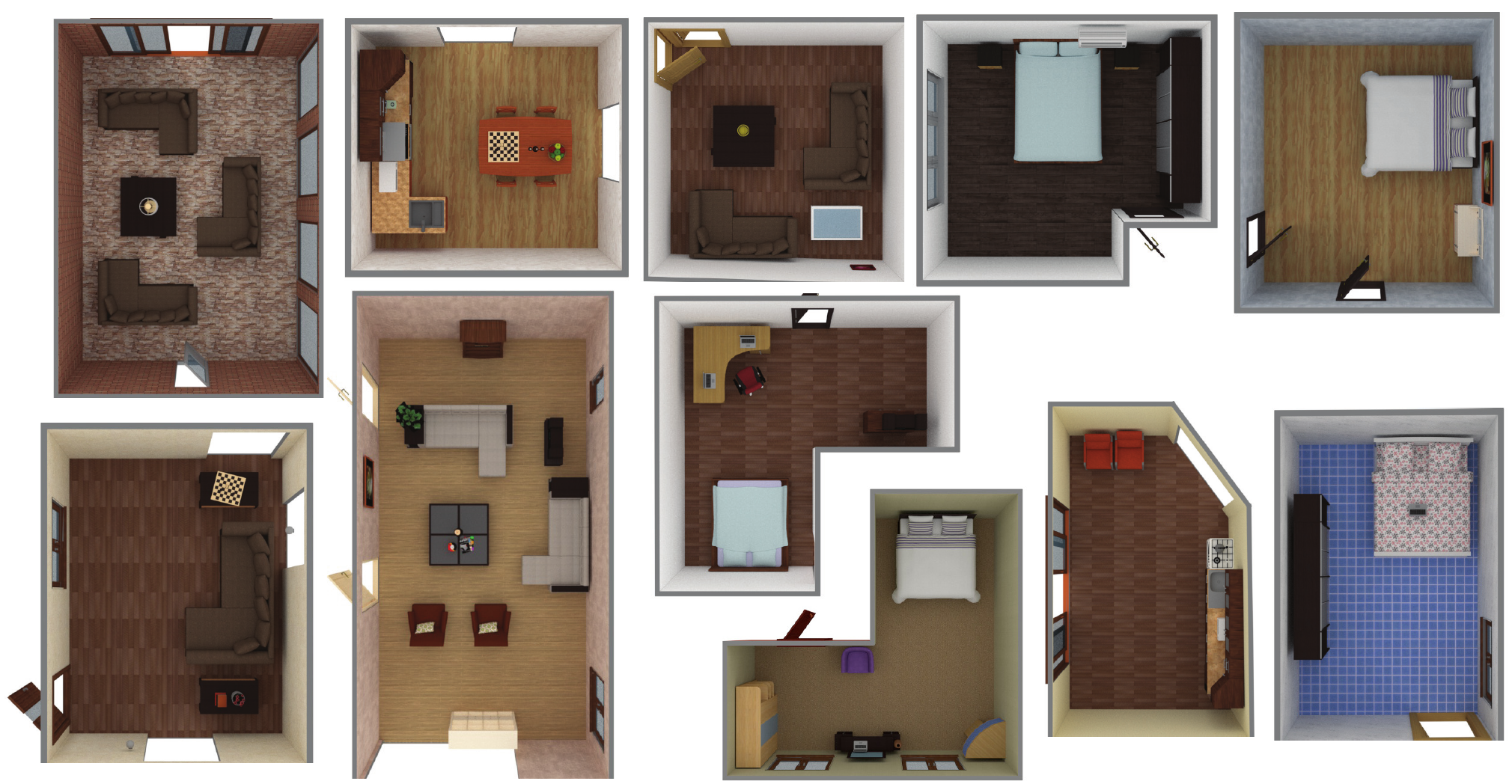}
    \vspace{-8pt}
    \caption{\label{fig:constr-samples-door} Samples from our model, with constraints. The sizes of the rooms, their shapes and door locations are constrained.}
\end{figure}

Figure~\ref{fig:constr-samples-door} shows sample layouts where the shapes of the rooms and the locations of doors where specified as constraints. Note that the doors are unobstructed in the generated layouts. 

\begin{figure*}[htbp]
    \centering
    \includegraphics[width=0.9\textwidth]{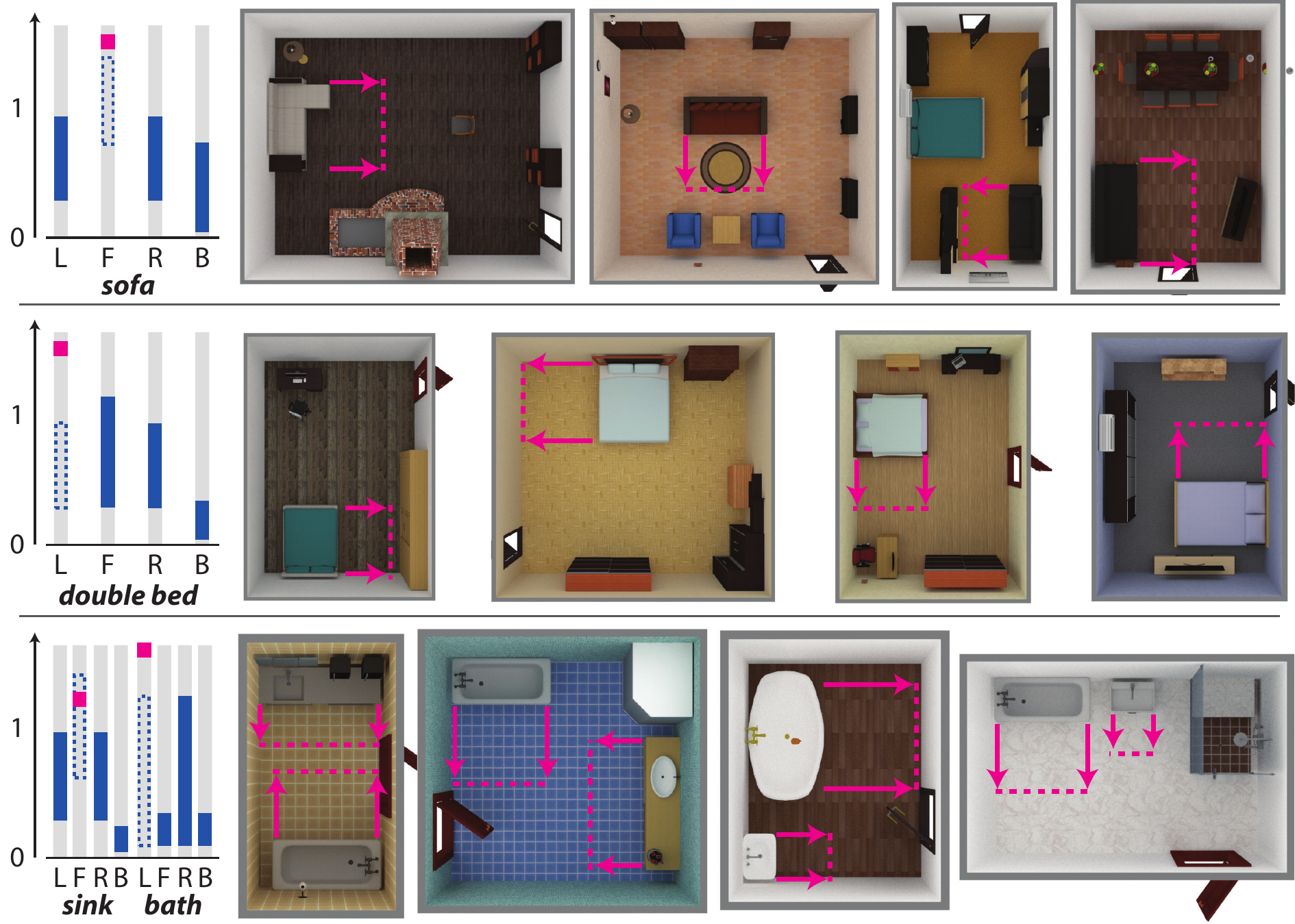}
    \caption{\label{fig:custom-padding-samples}Samples from our model, with user-specified clearance constraints. The left column shows the default (blue) and user-specified (pink) padding ranges in meters for each side (left/front/right/back) of the indicated object; the remaining columns show samples drawn from our model with the constraint applied, with the specified padding regions indicated}
\end{figure*}

Figure~\ref{fig:custom-padding-samples} shows examples of layouts where a user has specified particular clearances to be respected around specific object classes. 
The bar plots on the first column (solid blue) show the ranges of clearances learned during training on the left (L), front (F), right (R) and back (B) of the models sampled from four chosen classes (sofa, double bed, sink and bath).
The pink squares on the bar plots depict user modifications of the learned parameter (dashed blue rectangles). 
For each specified constraint (rows), four sample outputs are visualized (columns), and the impact of the user specification is shown using pink arrows. 
In this particular example, the constraints are imposed by directly editing learned parameters rather than using rejection sampling, which leads to faster runtime.

\begin{figure}
    \centering
    \includegraphics[width=\linewidth]{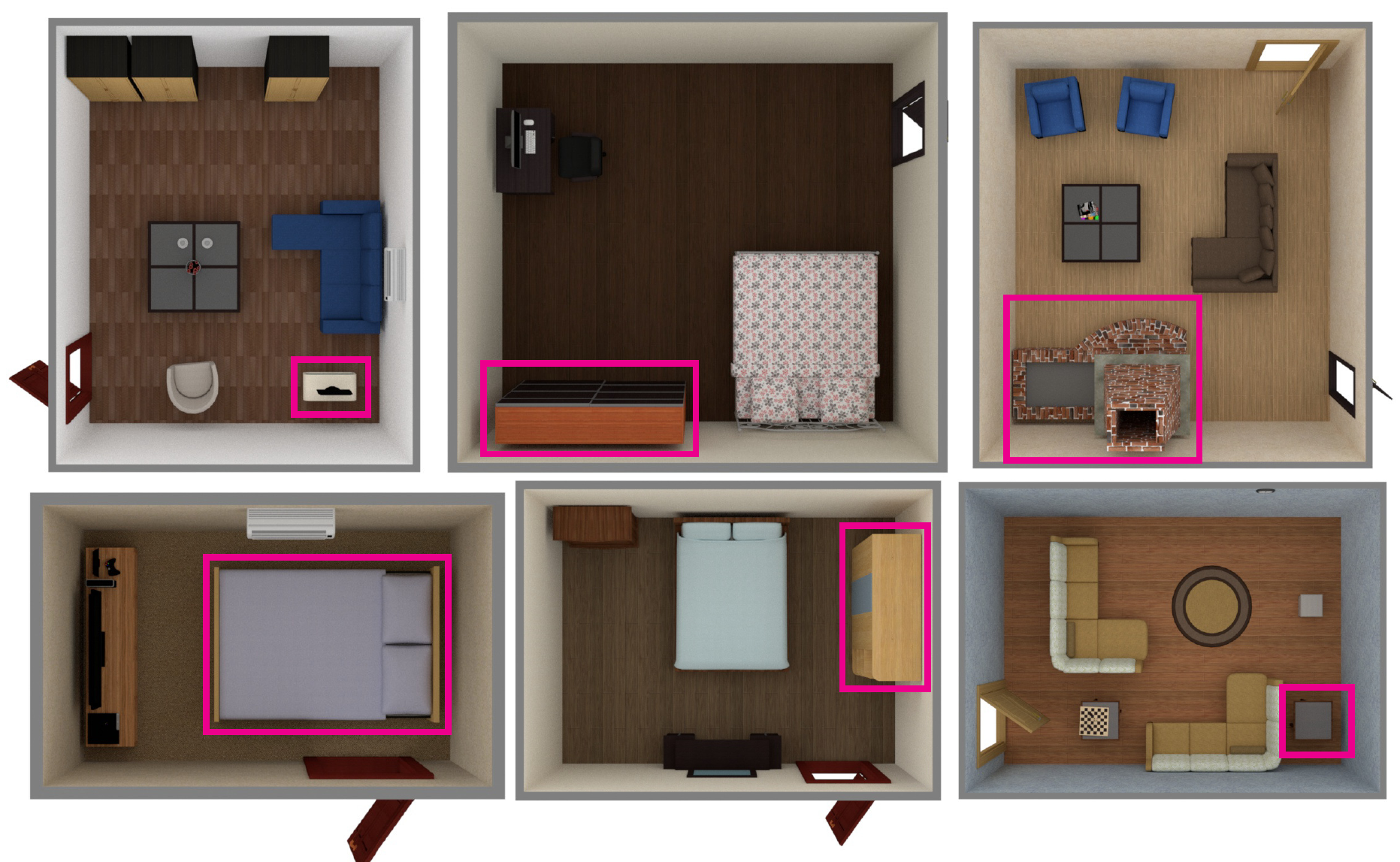}
    \caption{Samples from our model, applying constraints that are not satisfied by any layout in the training set. In each case, we constrain the room type, size, and placement of one object (indicated by a pink box), choosing a combination of constraints that is not satisfied by any layout in SUNCG. Our model is able to sample rooms fulfilling the constraints, despite not having seen such examples at training time.}
    \label{fig:ood-samples}
\end{figure}

\hdg{Constraints producing uncharacteristic layouts} 
One benefit of training a \textit{constrainable} generative model is that we can generate rooms fulfilling constraints that are never fulfilled in the training dataset, or only very rarely.
We demonstrate this by using random sets of reasonable constraints and identifying those sets of constraints which are not jointly satisfied by any room in the SUNCG dataset.
Then, we use our model to sample a room that does satisfy the constraints. Examples are given in Fig.~\ref{fig:ood-samples}.

\hdg{Runtime with constraints}
\tab{rejection-rates} shows the impact of applying different constraints on the running time of our method.
Our single-threaded Python implementation takes just 0.04s to sample an unconstrained room layout.
Even with complex constraints applied, the sampling time remains practical.

\begin{table}
    \centering
    \begin{tabular}{cc}
        \toprule
        \textbf{Constraint} & \textbf{Time per sample /s} \\
        \midrule
        unconstrained & 0.04 \\
        room type & 0.04 \\
        object class exclusion & 0.04 \\
        clearance & 0.04 \\
        traversability & 0.05 \\
        object placement & 1.4 \\
        gap placement & 1.8 \\
        room size & 6.8 \\
        size + doors + windows & 112 \\
        \bottomrule
    \end{tabular}
    \caption[Average time taken to sample a complete layout from our model, with different types of constraint applied]{
    Average time taken to sample a complete layout from our model, with different types of constraint applied. The timings are for an unoptimised Python implementation running on a single thread.
    }
    \label{tab:rejection-rates}
\end{table}

\subsection{User study} 
We assessed the realism of layouts generated using our model via a user study comparing its output to human-designed rooms from the SUNCG database.
We presented 1400 pairs of images to eight non-expert users and asked them to identify the image with a more realistic, or natural, layout of objects.
In each case, one image was a ground-truth (human-designed) layout from SUNCG, and the other was a sample from our model; the order of the two images was randomised for each pair.
%
The goal here is that our samples are indistinguishable from human-designed layouts, i.e.~of equal perceived quality to them, so users prefer ours 50\% of the time.

\begin{table}
    \centering
    \begin{subtable}{0.48\linewidth}
        \centering
        \begin{tabular}{cc}
            \toprule
            \textbf{Viewpoint} & \textbf{Ours pref.} \\
            \midrule
            overhead & $48.1 \pm 6.6\%$ \\
            1st person & $58.1 \pm 6.0\%$ \\
            \bottomrule
        \end{tabular}
        \caption{}\label{tab:user-study-unconstrained}
    \end{subtable}
    \begin{subtable}{0.5\linewidth}
        \centering
        \begin{tabular}{cc}
            \toprule
            \textbf{Constraints} & \textbf{Ours pref.} \\
            \midrule
            size + object & $45.2 \pm 6.8\%$ \\
            size + door & $35.2 \pm 5.4\%$ \\
            \bottomrule
        \end{tabular}
        \caption{}\label{tab:user-study-constrained}
    \end{subtable}
    \caption{
        Percentage of image-pairs where users preferred (i.e. deemed more realistic) a layout sampled from our model, as opposed to a ground-truth layout from SUNCG (`Ours pref.'). Higher is better, with 50\% indicating that our samples are indistinguishable from ground-truth. Ranges are the 95\% confidence interval~\cite{efron86}.
        \textbf{(a)} Unconstrained layouts;
        \textbf{(b)} Constrained layouts.        
    }
    \label{tab:user-study}
\end{table}

\hdg{Unconstrained}
We sampled several hundred random layouts from our model without constraints, and a similar number of ground-truth layouts from SUNCG.
We presented images in the form of either overhead renderings or first-person camera views from inside the room.
The observed user preferences are given in \tab{user-study-unconstrained}; we see that our layouts are equivalent in perceived quality to the human-designed layouts in the training set.
Specifically, in first-person views, users slightly preferred our layouts; in overhead views, our layouts are indistinguishable from ground-truth up to statistical significance.

\hdg{Constrained}
We assessed room layouts generated by our model with constraints as we did layouts without constraints, but using only overhead renderings.
We considered two representative settings for constrained generation:
(i) fixing the room size and placement of one object; and
(ii) fixing the room size and locations of doors and windows (implying gap placement and traversability constraints).
In both cases, we generated several hundred random examples.
For (i), we randomly generated arbitrary, but meaningful, pairs of constraints and sampled one layout fulfilling each.
For (ii), we randomly selected rooms from SUNCG, and used their size and door/window locations as constraints for our model, again sampling one layout for each.
In the second case, our model \textit{refurnishes} existing rooms.
In both cases, we compare our samples against arbitrary ground-truth rooms, which typically do not fulfill the same constraints,~i.e.~ we test the realism of our samples and not whether constraints are fulfilled (which is guaranteed by rejection).
Results are given in \tab{user-study-constrained}.
With room size and the placement of one object constrained, our layouts are indistinguishable from ground truth up to statistical significance.
With room size and the positions of doors constrained, users preferred human-designed layouts.

\begin{table}
    \centering
    \begin{tabular}{cc}
        \toprule
        \textbf{Room type} & \textbf{Ours pref.} \\
        \midrule
        bedroom & $75.0 \pm 4.9\%$ \\
        living room & $71.1 \pm 5.6\%$ \\
        \bottomrule
    \end{tabular}
    \caption{Comparison with Wang et al~\shortcite{wang18tog}: percentage of image-pairs where users preferred a layout sampled from our model, as opposed to one generated by \shortcite{wang18tog}.}
    \label{tab:wang-study}
\end{table}

\hdg{Comparison with \cite{wang18tog}}
We compared randomly generated unconstrained samples from our model with those generated by the state-of-the-art CNN-based method of \cite{wang18tog}.
We presented users with 550 pairs of images, each showing one layout generated by our method and one by that of \cite{wang18tog}.
We restricted the room types to those supported by both our method and theirs, i.e.~bedroom and living room.
The users were again asked to identify the image with a more realistic arrangement of objects.
Results from this experiment are shown in \tab{wang-study}.
We see that our layouts are often preferred to those of \cite{wang18tog}.
Moreover, our model is interpretable and fast to train, whereas theirs is a non-interpretable black-box model trained over several days.

\section{Discussion} \label{sec:discussion}

\hdg{Comparison with prior works}
Probabilistic generative methods for room layouts are challenging to sample from. Often the sampling is not guaranteed to converge to a valid layout, espcially when many objects are present.~e.g.~
the model proposed by Handa et al~\shortcite{handa16icra}. This particular model also requires that the number of objects, and size of the room, be specified manually.
Our model performs favorably compared with with the very recent work~\cite{wang18tog} that learns millions of parameters over days of training.
For over 70\% of pairs presented, users preferred our layouts to theirs.
In addition to accommodating constraints easily, our model has another advantage in that the parameters learned are over semantically meaningful concepts (categories such as furniture) allowing direct modulation of learned parameters as shown in Figure~\ref{fig:custom-padding-samples}. Although we manually specified padding constraints, they could be calculated from alternatives such as human-centric affordances~\cite{qi18cvpr}.

\begin{figure}
    \centering
    \includegraphics[width=\linewidth]{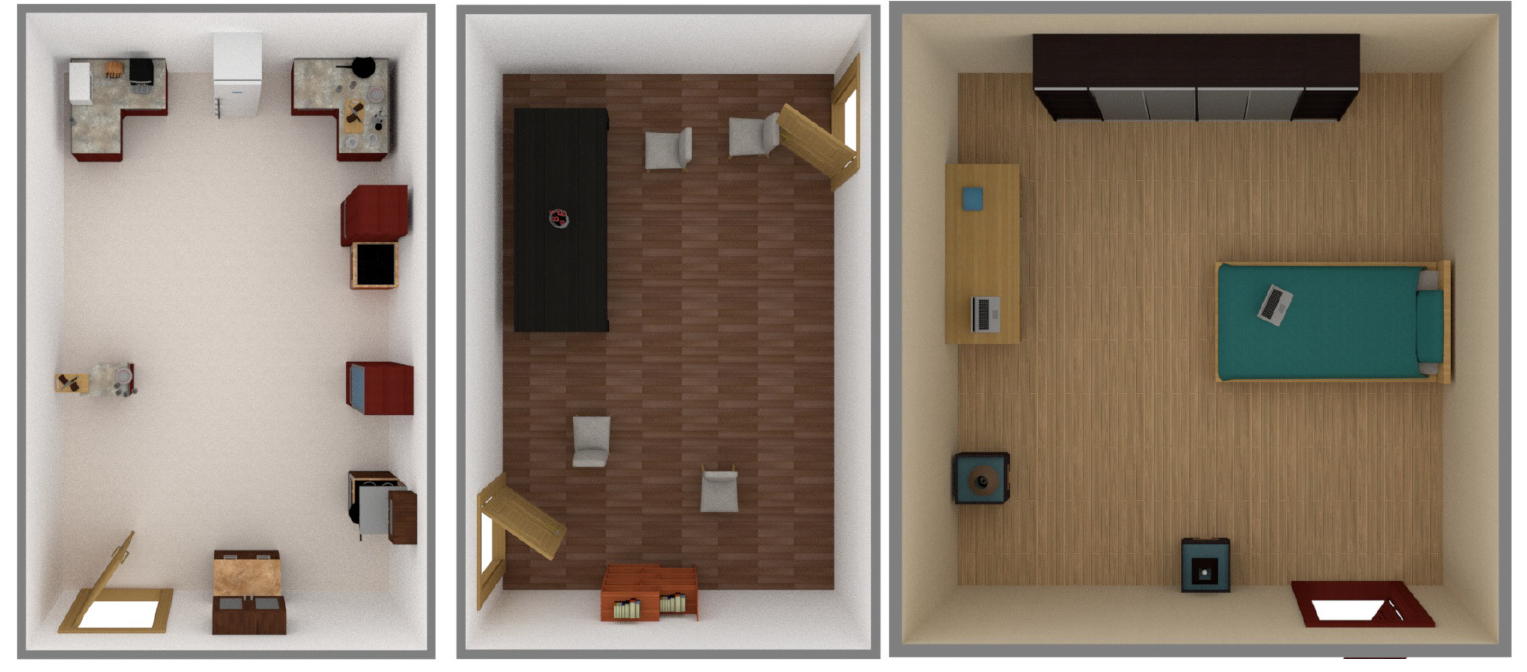}
    \caption{Samples from our model, but without motifs and abutments. Left: the kitchen cabinets and appliances are scattered, rather than placed adjacent to one another (as enabled by abutments). Centre: the chairs are scattered, rather than placed around the dining table (as enabled by motifs). Right: the two night-stands (lower left) are not at the expected location near the bed (as enabled by motifs)}
    \label{fig:ablated-samples}
\end{figure}

\hdg{Inter-object relationships}
We explicitly discover and encode relationships across classes of objects using patterns such as motifs and abutments. These patterns capture higher order relationships (not just pairwise); in the case of abutments, they are able to model sequences of variable lengths which may not be present in the training database. Figure~\ref{fig:ablated-samples} shows unnatural layouts generated when inter-object relationships due to motifs and abutments are not modeled. Additionally, implicit relationships are captured between different CAD models of the same class in a given layout, through the conditioning on the number $n_c$ of objects already placed (line 5 of algorithm~\ref{alg:sampfurniture}). For example, the generative process may not favor a large item from a class if multiple small items from that class have already been sampled.

\hdg{Novelty of samples}
Large generative models run the risk of over-fitting their training set, memorizing the training data instead of generalizing to produce novel samples.
Fig.~\ref{fig:ood-samples} shows that our model is able to generate constrained layouts even when the constraints are not fulfilled by any room in the training set.
Thus, it is not simply memorizing the training data, but is creating new arrangements.
As a second demonstration of this, we directly searched for the most similar layouts in the training set, to layouts generated by our model.
The retrieved layouts are typically quite different from our samples in arrangement and exact object counts, which confirms that our model is generalizing.

\begin{figure}
    \begin{center}
        \begin{tabular}{@{}c@{~~~~}c@{~~~~}c}
            \begin{tabular}{@{\hspace{-8pt}}c@{}c@{}}
                \includegraphics[width=.15\linewidth]{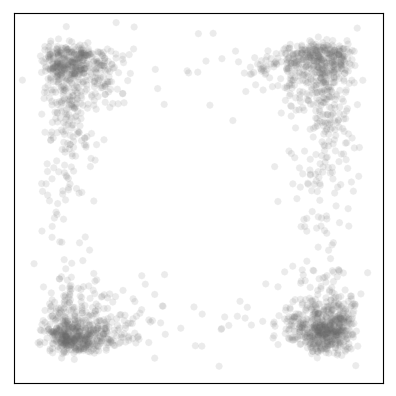} &
                \includegraphics[width=.15\linewidth]{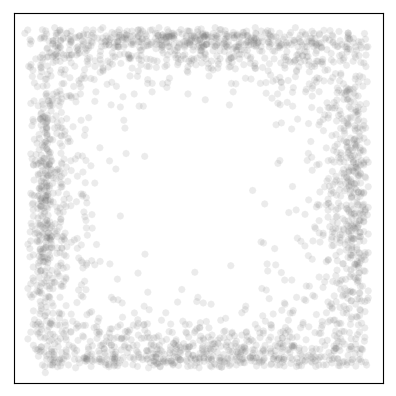} \\
                \includegraphics[width=.15\linewidth]{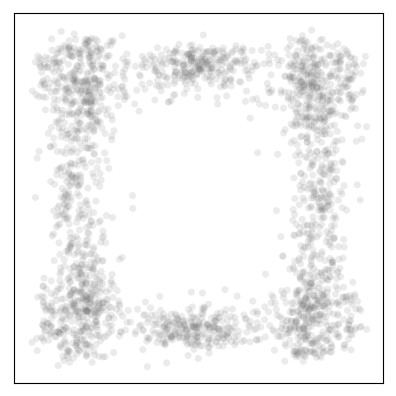}&
                \includegraphics[width=.15\linewidth]{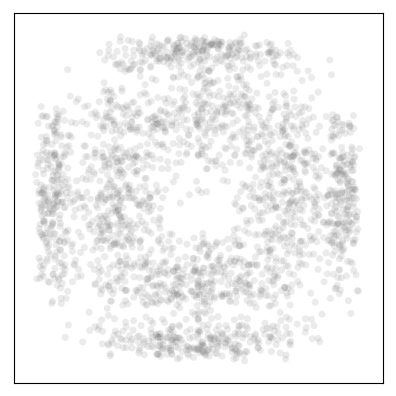} \\
                \includegraphics[width=.15\linewidth]{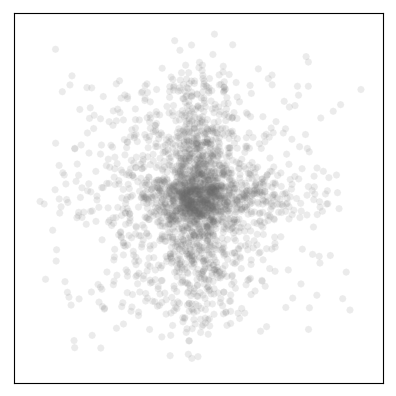}&
                \includegraphics[width=.15\linewidth]{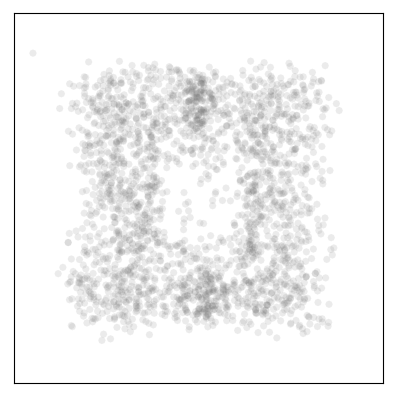}
            \end{tabular}
            &
            \begin{tabular}{@{}c@{~~}c@{}}
                \toprule
                \textbf{furniture} & $p_{\mathrm{edge}}$ \\
                \midrule
                toilet & 0.88 \\
                range oven & 0.86 \\
                sink & 0.83 \\
                \midrule
                chair & 0.32 \\
                tripod & 0.32 \\
                armchair & 0.29 \\
                \bottomrule
            \end{tabular}
            &
            \begin{tabular}{@{}c@{~~}c@{}}
                \toprule
                \textbf{furniture} & $p_{\pi/2}$ \\
                \midrule
                kitchen cabinet & 0.99 \\
                dishwasher & 0.99 \\
                single bed & 0.99 \\
                \midrule
                office chair & 0.74 \\
                armchair & 0.65 \\
                tripod & 0.49 \\
                \bottomrule
            \end{tabular}
            \\
            (a) & (b) & (c)
        \end{tabular}
        \caption{\label{fig:interpretable} Many parameters that are learned during training are human-interpretable.
        \textbf{(a)} Heat-maps showing locations where our model places different objects. Clockwise from top-left: shower, cabinet, sofa, double bed, dining table and toilet.
        \textbf{(b)} furniture classes with highest (top) and lowest (bottom) probability $p_{\mathrm{edge}}$ of being at the edge of a room rather than the interior
        \textbf{(c)} furniture classes highest (top) and lowest (bottom) probability $p_{\pi/2}$ of being at an angle that is a multiple of $\pi/2$.
        }
    \end{center}
\end{figure}

\hdg{Efficient implementation of constraints}
For many of the constraints listed in Section~\ref{sec:constraints}, rejection sampling can be avoided using alternative implementations. For example, space constraints may be tailored at the class level by biasing the 4D Normal distribution learned for padding. Figure~\ref{fig:custom-padding-samples} shows direct editability of learned parameters. Example layouts produced by the modified distribution are shown on the right, along with the effects of the user manipulation on the corresponding objects. Another example of a constraint that may be implemented efficiently is the specification of certain object classes (or CAD models) as not desirable. Rather than relying on rejection sampling, these classes (or models) may be pre-emptively avoided during instantiation.

\hdg{Interpretability} 
Since our model learns parameters associated with semantically meaningful categories and positions, the learned results can be interpreted and manipulated. 
Figure~\ref{fig:interpretable}a visualizes heat-maps of where the model places a few chosen object classes. 
For each class, we sampled 2500 rooms, and plotted (black dots) where objects of the indicated class were placed (normalizing the room bounds to a unit square).
The model has learnt to place different classes meaningfully -- for example, showers are almost always at the corner of a bathroom, dining tables are often at the center of a room, and toilets are always against a wall. Figure~\ref{fig:interpretable}b and Figure~\ref{fig:interpretable}c list the highest and lowest probability entries for object positioning and orientation.
We obtained these numbers by averaging over the corresponding probabilities for all CAD models in the stated classes. The numbers align with our expectation that chairs and tripods are not typically placed along the edges of rooms, and that they are less likely to be aligned with the walls than beds, kitchen cabinets or dishwashers.

\hdg{Room shape}
In our implementation we decompose non-rectangular rooms into rectangular regions and apply our cell structure on each region. 
Other strategies to partition rooms into cells may be adopted as long as they are kept consistent across training and sample generation. 
However, the choice of partitioning strategy may impact the quality of results. 

\hdg{Multiple, simultaneous constraints} 
Another advantage of rejection sampling as a general mechanism to impose constraints is that support for multiple constraints is trivial to implement.
However this flexibility is accompanied by a penalty in terms of runtime. 
The time taken to generate a sample that satisfies multiple constraints, on average, is the product of the times taken to support each constraint.

\hdg{Limitation: a posteriori editing} 
Our model is designed for fast generation of layouts with pre-specified constraints but our formulation does not facilitate editing an existing (previously sampled) layout.
While we can handle a priori specification such as ``I want a new layout such that this television set is placed against the east wall'', it cannot handle a posteriori editing such as ``In the previously generated layout, move the television to the east wall''.

\section{Conclusion}
We have presented an efficient, probabilistic, data-driven, generative model for indoor furniture layouts.
The algorithm used to generate layouts is simple and the parameters learned from training data are human-interpretable. 
We demonstrated that our model is able to accommodate a variety of constraints using rejection sampling as well as editing of learned parameters. 
We presented qualitative and quantitative results through rendered layouts, performance measurements and a user study.
These showed that our layouts are realistic, and preferred by users over the state-of-the-art method of \cite{wang18tog}.

\bibliographystyle{eg-alpha-doi}
\bibliography{./bibtex/shortstrings,./bibtex/calvin,./bibtex/vggroup,./bibtex/sigga18}

\newpage

\end{document}